\documentclass[letterpaper, 10 pt, conference]{ieeeconf}  

\usepackage{graphicx}
\usepackage{latexsym}
\usepackage{booktabs}
\usepackage{multirow}
\usepackage{amssymb}
\usepackage{amsfonts} 
\usepackage{algorithm}
\usepackage{algorithmic}
\usepackage{amsmath}

\usepackage{amsthm}
\usepackage{times}
\usepackage{soul}
\usepackage{url}
\usepackage{bm}

\IEEEoverridecommandlockouts                              
\title{\LARGE \bf
Navigating the Smog: A Cooperative Multi-Agent RL for Accurate Air Pollution Mapping through Data Assimilation
}

\author{Ichrak Mokhtari$^{1}$,  Walid Bechkit$^{1}$, Mohamed Sami Assenine $^{1}$ and Hervé Rivano$^{1}$ 
\thanks{*This work was supported by the French National Research Agency (ANR) under Project ANR-21-CE25-0003 (DRONMAP)}
\thanks{$^{1}$Ichrak Mokhtari,  Walid Bechkit, Mohamed Sami Assenine and Hervé Rivano are with INSA Lyon, Inria, CITI, UR3720, 69621 Villeurbanne, France
        {\tt\small ichrak.mokhtari@insa-lyon.fr, \newline 
        walid.bechkit@inria.fr, \newline
        mohamed-sami.assenine@inria.fr,\newline herve.rivano@inria.fr}}%
}

\begin{document}

\maketitle
\thispagestyle{empty}
\pagestyle{empty}

\begin{abstract}

The rapid rise of air pollution events necessitates accurate, real-time monitoring for informed mitigation strategies. 
Data Assimilation (DA) methods provide promising solutions, but their effectiveness hinges heavily on optimal measurement locations. This paper presents a novel approach for air quality mapping where autonomous drones, guided by a collaborative multi-agent reinforcement learning (MARL) framework, act as airborne detectives. Ditching the limitations of static sensor networks, the drones engage in a synergistic interaction, adapting their flight paths in real time to gather optimal data for Data Assimilation (DA). Our approach employs a tailored reward function with dynamic credit assignment, enabling drones to prioritize informative measurements without requiring unavailable ground truth data, making it practical for real-world deployments.
Extensive experiments using a real-world dataset demonstrate that our solution achieves significantly improved pollution estimates, even with limited drone resources or limited prior knowledge of the pollution plume. Beyond air quality, this solution unlocks possibilities for tackling diverse environmental challenges like wildfire detection and management through scalable and autonomous drone cooperation.

\end{abstract}

\section{INTRODUCTION}

All over the world, air pollution is becoming a severe issue, given its acute effects on human health and the environment. To better understand the pollution origins and reduce its negative impacts, it is of great importance to develop effective monitoring strategies that enable the generation of accurate pollution map estimates. These maps would allow decision-makers to take adequate actions on air pollution reduction, industrial control, and land use management. Moreover, the public may be more aware of their air quality, which would encourage them to act more responsibly and contribute to the collective ecological effort.

Traditionally, air pollution monitoring is performed using fixed air quality monitoring stations, placed sparsely in urban areas. Although highly precise, these stations are inflexible, expensive, and provide poor spatial granularity. Therefore, interpolation methods fail most of the time to estimate pollution concentrations in unmeasured areas and to produce representative accurate pollution maps at low scales, although some pollutants are known to considerably fluctuate at very small scales. Indeed, most research studies using measurements from monitoring stations provide estimations at a city scale or higher \cite{liang2023airformer,gao2019incomplete,du2019deep}. 

With the development of low-cost micro-scale sensing technology in the last decade, the use of wireless sensor networks has become widespread, hence providing a flexible and viable alternative to conventional monitoring stations. Furthermore, the deployment of these sensing devices on mobile platforms like Unmanned Aerial Vehicles (UAVs), also known as drones, offers a new impetus to air pollution monitoring, allowing measurements in hitherto unreachable areas, and providing wider coverage.

On the other side, the integration of sensor measurements into mathematical models through methods such as Data Assimilation (DA) allows for a spatially continuous representation of pollution concentrations with greatly reduced bias. Indeed, coupling effectively real observations with numerical simulations to reduce the inherent uncertainty of both sources enables better air pollution mapping by constraining physical models with observations and allowing better estimations. Therefore, data assimilation methodologies are by far one of the most versatile and used approaches in earth's dynamical systems and particularly in the air quality field \cite{nguyen2021data,filoche2020completing}. Nevertheless, DA quality is strongly affected by measurement locations.

In this paper, we investigate the autonomous drones path planning problem for an effective data assimilation of air pollution observations. Here, drones are tasked to find the sequence of optimal sensing positions permitting to improve the best data assimilation estimates. Many challenges are encountered when facing this problem, which we summarize in the following points. Firstly, given that ground truth data are ignored in practice, drones trajectories guidance is particularly complicated. Secondly, the combined contribution of all observations on the quality of air pollution mapping through data assimilation cannot be adequately expressed based on individual contributions of each observation. Furthermore, the selection of some sets of sensing locations may deteriorate the previous estimates. Finally, the joint action space grows significantly with the number of drones and their possible actions.

In the past decade, Reinforcement Learning (RL) \cite{sutton1998introduction} has emerged as a viable and powerful approach to yield optimized policies for sequential decision-making problems. In our specific context, the complex and dynamic decision-making process, the huge research space encountered and the substantial impact of sensing location on data assimilation quality, makes our problem suitable to be tackled by RL. 
Recent years have witnessed the rise of Deep Reinforcement Learning (DRL), leveraging deep neural networks as function approximators, particularly in robotics applications such as drone path planning. For example, Theile et al. applied DRL based on double deep Q-network \cite{van2016deep} and proved RL's effectiveness in constraints coverage path planning task for UAVs \cite{theile2020uav}. On another track, Jing et al. proposed a near-optimal trajectory for autonomous drone racing based on an RL framework \cite{song2021autonomous}. 

Although numerous works have focused on UAVs path planning, only a limited number have explored the design of a multi-drone system based on DRL \cite{yun2022cooperative}. Furthermore, most existing works are designed with different goals than the one treated here. For instance, in \cite{assenine2023cooperative}, a fleet of drones is considered to assess pollution plumes modeled as Gaussian Processes. This work's main goal is to characterize the parameters of Gaussian processes using a DRL framework. In contrast, we focus on improving pollution maps through enhancing the quality of data assimilation.

In this paper, we present an efficient Multi-Agent RL (MARL) strategy to select at each step the optimal drones positions for air pollution mapping through data assimilation. To tackle the problem of joint action explosion and achieve agents' cooperation at the same time, we develop an independent Q-learning strategy based on credit assignment, in which each UAV agent learns its own policy and has its own reward deduced from a common team gain while sharing the global state with others agents. The main contribution of this paper can be summarized as follows:

\begin{itemize}
   \item  As far as we are aware, this is the first attempt to use cooperative multi-agent reinforcement learning in the field of data assimilation for air pollution mapping. Given raw estimated pollution concentrations and UAVs positions as the state description, our agents maximize the improvement of data assimilation outputs.
   
   \item We derive a team reward function that does not rely on ground truth data, permitting to improve data assimilation estimates over time. Furthermore, two credit assignment schemes were investigated in this specific context.

   \item We conduct extensive experiments using a real-world dataset. The results show that our proposal effectively improves data assimilation quality and leads to accurate estimated pollution maps. Furthermore, we show that the proposed MARL strategy is effective even for poor initial simulations.
\end{itemize}

\section{Background}

In this section, we briefly review the background of data assimilation as well as one of its most used methods. Then, we go over some preliminaries of multi-agent RL.

\subsection{Data Assimilation}
Data Assimilation (DA) methods have been widely identified as effective techniques for weather forecasting in general and increasingly for air pollution mapping in recent years \cite{lahoz2010data,reich2015probabilistic}. Specifically, the objective of DA approaches is to estimate the true state of a system (unknown by definition), denoted by a vector $x^{t}  \in \mathbb{R}^{n \times 1}$ ($t$ refers to truth) using a priori estimate of the system state $x^{b} \in \mathbb{R}^{n \times 1}$ (b stands for background) and real measurements vector $y \in \mathbb{R}^{m \times 1}, (m<n)$. These observations may be performed in any location and are in practice sparse. In DA solutions, observations are used to reduce errors committed in the initial estimate and the output can be given by :
\begin{equation}
    x^{a} = x^{b} +\nabla x
\end{equation}

where $x^{a} \in \mathbb{R}^{n \times 1}$ represents the DA result referred to as analysis, and $\nabla x$ denotes the correction of the background state $x^{b}$. This correction depends on the DA technique considered and is performed in a way that $x^{a}$ approaches at best $x^{t}$.

\subsection{Best Linear Unbiased Estimator}
Best Linear Unbiased Estimator (BLUE) is one of the most used data assimilation methods \cite{nguyen2021data,tilloy2013blue}. It aims to estimate the true state of a system $x^{t}$, assuming unbiased background errors $\epsilon^{b} $ and unbiased observation errors $\epsilon^{m}$. Furthermore,  $\epsilon^{b}$ is assumed to be uncorrelated with $\epsilon^{m}$. The estimate given by BLUE is formulated as a linear combination of $x^{b}$ and $y$ and is given by the following equation: 

\begin{equation}
      x^{a} = x^{b} + K( y - Hx^{b} )
       \label{eq:1}  
\end{equation}

where $K \in \mathbb{R}^{n \times m}$ is the Kalman gain and $H \in \mathbb{R}^{m \times n}$ the observation operator. Concretely $H$ is a binary matrix where $h_{ij}$ = 1 if a sensor $i$ is deployed 
at position $j$, and $h_{ij}$ = 0 otherwise. From expression \ref{eq:1},  the correction of the background $x^{b}$ is performed using $K(y - Hx^{b})$. Given the objective of minimizing the trace of the covariance matrix of the analysis errors, it is proven that the optimal Kalman gain is given by:
\begin{equation}
       K = BH^{T}(HBH^{T} +R)^{-1}
\end{equation}
with $B$ and $R$ representing the covariance matrices of background errors and observation errors, respectively.

\subsection{Multi-Agent Reinforcement Learning}

Multi-Agent Reinforcement Learning (MARL) is an up-to-date artificial intelligence paradigm for modeling  Multi-Agent Systems (MASs) \cite{panait2005cooperative}, that has proven to be extremely effective in a wide range of problems including robotics applications \cite{wei2021multi,yun2022cooperative,peng2021facmac}. Within this configuration, a set of agents learn to optimize their policies through interaction with the environment while considering at the same time other agents' behaviours. Usually, MARL problems are formulated under Markov Game (MG) \cite{littman1994markov}, a framework that extends the standard Markov Decision Process (MDP) used for single-agent RL problems.

Cooperative learning is one important MARL scheme as different real-world's problems can be formulated as distributed systems with multiple agents collaborating to achieve a common objective. As stated by \cite{panait2005cooperative}, there are two main configurations to deal with cooperative MARL namely team learning and concurrent learning. In the former case, a single-agent RL is used to model the whole system and works out the behaviour of the team, using the joint action instead of local actions. Although simple as it implies using standard RL algorithms, it poses scalability issues, reflecting the exponential growth of the state action space with the increased number of agents. On the other extreme opposite, concurrent learning also called independent learning, copes with scalability problems, as it considers a fully decentralized scheme with agents maximizing their own rewards based on local observations. Therefore, each agent has its own learning process to adjust its behaviour. Nonetheless, this conventional independent learning strategy rules out the cooperation between agents and ignores the resulting non-stationarity issue.

Independent learners can enhance their team performance by sharing some information with other agents (e.g. their observed states) and ensuring cooperation by splitting a team reward among individual agents. This split is known as the credit assignment problem, which remains to this day a challenging research topic in MARL. Intuitively, if an agent's gain is insufficient to compensate for its efforts, the agent may be deterred from cooperating or may become a freeloader. Some literature research has reported that an equal reward split known as global reward can lead to promising results if a good exploration strategy is chosen \cite{claus1998dynamics}, beating in some cases credit assignment based on local reward (i.e., agents are rewarded based solely on their individual behavior) \cite{balch1997learning}. Other studies demonstrate experimentally that difference-rewards based credit assignment achieves better results than global reward \cite{wei2021multi}. The difference rewards method considers that the reward to an individual agent is determined as the difference between the team reward from the joint action and that by deleting the agent's action.
%


\section{DRL-based UAVs Path Planning for Data Assimilation} 
In this section, we present the UAVs path planning problem for data assimilation. We model our system by markov game framework and we present our MARL learning scheme.

\subsection{Problem Description}
We consider a map discretized in a set of points $  \mathcal{P} $ that approximate a given region to be monitored with a high spatial granularity $\mid \mathcal{P} \mid= n $. Within this area, we consider a fleet of autonomous UAVs of size $N$ denoted $\mathcal{U}= \{u_{1}, u_{2}, ..., u_{N}\}$, moving between the $n$ potential positions. Specifically, we assume a deployment of rotary-wings UAVs, a type of drones capable of performing hovering and stabilizing when sensing. These UAVs are equipped with adequate sensors, permitting their localization, measurements sensing, and connectivity to a leader UAV or a ground station. 

Each drone location at time $t$ is given by its coordinates $(x^{u_{i}}_{t}, y^{u_{i}}_{t}, h^{u_{i}}_{t}) \in \mathbb{R}^{3}$, where $x^{u_{i}}_{t}$, $y^{u_{i}}_{t}$ and $h^{u_{i}}_{t}$ represents its latitude, longitude and altitude. To avoid collisions, drones are assumed to fly at slightly different altitudes. Furthermore, UAVs are limited to moving with constant velocity $V$ or hovering, thus a drone's velocity at time $t$ is given by  $v^{u_{i}}_{t} \in \{0, V \}$.
The time horizon is divided into slots $\bigtriangleup t$ representing the maximum task duration of UAVs depending on their trajectory length. During the latter, all drones move to their new positions and perform hovering to sense the environment. Therefore, the length of a given time slot is given by $\bigtriangleup t =  \underset{u_{i} \in \mathcal{U}} {max} \{(t^{u_{i}}_{movement} + t^{u_{i}}_{hovering})$ \}.
Each drone $u_{i}$ has its own budget $b^{u_{i}}_{t}$ that evolves over time. This budget is related to drones' battery capacity and is gradually consumed mainly by the mobility of drones, which has a cost proportional to the distance traveled. For the sake of simplicity, we don't consider the hovering cost of UAVs after moving to their new positions. Furthermore, each drone mission is considered over whenever its budget is completely consumed.

\begin{figure}[h]
\hspace{-0.3cm}
\includegraphics[scale=0.49]{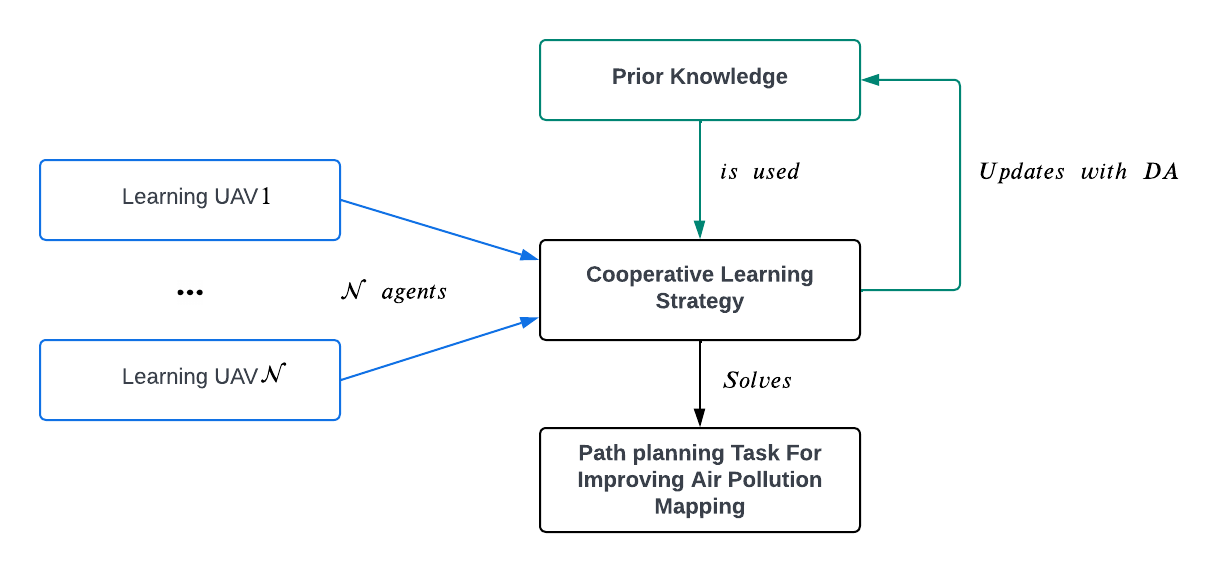}

\caption{Illustration of the general data assimilation problem using a cooperative fleet of UAVs.}
\label{fig:caract}
\end{figure}

With that said, our problem can be formulated as follows. Given an initial background $x^{b}$, that may be represented by a simulated pollution map, a set of UAVs to be sequentially redeployed in the monitored region $\mathcal{P}$, we aim to find the UAVs optimal paths (sequence of sensing positions), permitting at each step to better improve the pollution map estimate (the analysis) $x^{a}$ given by data assimilation, using the set of the collected observations $y$, so that we approach as much as possible the ground truth $x^{t}$. In other words, our goal consists of finding a drones cooperative path planning strategy allowing an efficient correction of data assimilation outputs over the steps, while considering at the same time the limited battery lifetime constraints of drones. Figure \ref{fig:caract} illustrates an overview of the path planning problem for data assimilation correction.

One of the main difficulties regarding optimizing mobile sensors positions for data assimilation correction is that the improvement of previous estimates is not guaranteed. Indeed, some sensing positions may deteriorate the previous estimate instead of improving it. Furthermore, the drones' trajectories can not be guided following the ground truth as the latter is ignored in practice. Additionally, the short flying duration of UAVs limits their missions and hence reduces the number of collected observations over time. On another hand, the research space at each step can become exceedingly large, following the size of the monitored region and the number of mobile sensors (UAVs), which poses a significant challenge to path planning. 
Thus, developing an efficient strategy for drones' trajectories that balances the trade-off between exploring new regions and exploiting existing information while avoiding worsening last estimates is crucial. All these put together motivates the use of RL to handle the combinatorial nature of the problem in such a complex environment by enabling efficient exploration-exploitation of the available solutions. As far as we are aware, this is the first time the optimization of UAVs paths to better assimilate data in physico-chemical models is considered. While this problem is specifically presented in the context of air pollution mapping, its significance extends beyond this application, given the widespread use of data assimilation in a range of fields such as weather forecasting, climate modeling, geophysics, and robotics \cite{asch2016data}.

\subsection{Model Framework as Markov Games}

We consider a multi-UAV coexisting environment, in which a set $\mathcal{U}$ of autonomous UAVs fulfill a collaborative task, consisting of correction continuously data assimilation outputs given an initial background. We define the corresponding MG framework by the tuple
 $(\mathcal{U}, \mathcal{S}, \mathcal{A}, \mathcal{P}, \mathcal{R}, \gamma)$ and we define its necessary elements in what follows.

\subsubsection{Set of Agents}
From the perspective of DRL, each UAV $u_{i} \in \mathcal{U}$ is modeled as a learning agent, which has the capability of sensing pollution measurements and selecting the next locations to fulfill an objective shared with other agents. 
 
\subsubsection{State}

The global state at time instant $t$ in the discretized map environment of size $n$ is given by $ s_{t} = ( p_{t}, {b_{t}}, x^{a}_{t} )$ and consists of the following components:
\begin{itemize}
    \item $\mathrm{\bm{p_{t}}} = \{(x_{t}^{u_{i}}, y_{t}^{u_{i}} ) \in \mathbb{R}^{2}, u_{i} \in \mathcal{U}\}$ represents the UAVs 2D positions at time $t$. UAVs' heights may be considered in the case of 3D pollution map construction.
 
    \item $\mathrm{\bm{b_{t}} =\{b_{t}^{u_{i}} \in \mathbb{R}, u_{i} \in \mathcal{U}\}  }$ corresponds to the UAVs remaining flying budgets. The initial available budget for a drone path, denoted $B_{max}$ may differ between drones. After each drone's movement, its budget $b_{t}^{u_{i}}$ is updated following the cost of the action taken:
    \begin{equation}
           b_{t}^{u_{i}} = b_{t-1}^{u_{i}} - c((x_{t-1}^{u_{i}}, y_{t-1}^{u_{i}}), (x_{t}^{u_{i}}, y_{t}^{u_{i}})) 
    \end{equation}

    \item $x^{a}_{t} \in \mathbb{R}^{n}$ 
is the estimate of the pollution map at time $t$ given by data assimilation.
\end{itemize}

 \subsubsection{Action}

The main decision of an agent is moving to one of the possible $n$ positions of the monitored region. Therefore, the joint action space of the set of ${N}$ UAVs is given by:
 \[ \mathcal{A} = \mathcal{A}^{u_{1}} \times ...\times \mathcal{A}^{u_{N}} = \{1, 2, 3, ..., n\}^{N} \]
 
 Each time a drone selects an action and moves to its new position, it performs hovering and measures pollution concentration. When a drone's budget is depleted, it lands and disappears from the monitored region. Hence, a dummy action is added in this case, corresponding to the action of landing. 

\subsubsection{State Transition}
At each time slot, after all agents decide on their actions, all the collected observations are used together to update the last estimate of the pollution map through data assimilation. The drones' budgets are adjusted and the global state is updated.

 \subsubsection{Reward Function}
 
In this work, we set a reward function to improve the outputs of the data assimilation method $x^{a}_{t}$ and accelerate the convergence of the latter to the ground truth map $x^{t}$ while attempting to use drones flying budgets wisely. We define a team reward as the mean difference error between two consecutive estimated pollution maps during a time step. Equation (\ref{eq:rt_team}) shows its mathematical expression for the case of overestimated simulations. Mainly,
our intuition stems from two observations: 1) numerical model simulations often overestimate reality \cite{soulhac2017model}, and 2) data assimilation methods improve the estimates on average and allow convergence to the ground truth map in the long run. Therefore, our reward can be interpreted as the mean convergence speed of data assimilation during a small time lap. The greater its value, the more impactful the agents' actions are to the improvement of the previous estimate. On the contrary, negative rewards imply a deterioration of the past analysis. Therefore, drones are encouraged to select positions that maximize the correction of the data assimilation at each step. 

\begin{equation}
   R_{t}=  r(s_{t}, a_{t}) =  \mathbb{E}[x^{a}_{t-1} - x^{a}_{t}]
    \label{eq:rt_team}
\end{equation}
While our primary focus is on the more common scenario of overestimated simulations, the reward function can be easily adapted to handle underestimations. In such cases, the reward would simply be the negation of Equation (\ref{eq:rt_team}).

\begin{figure*}[h]
\centering
\includegraphics[scale=0.5]{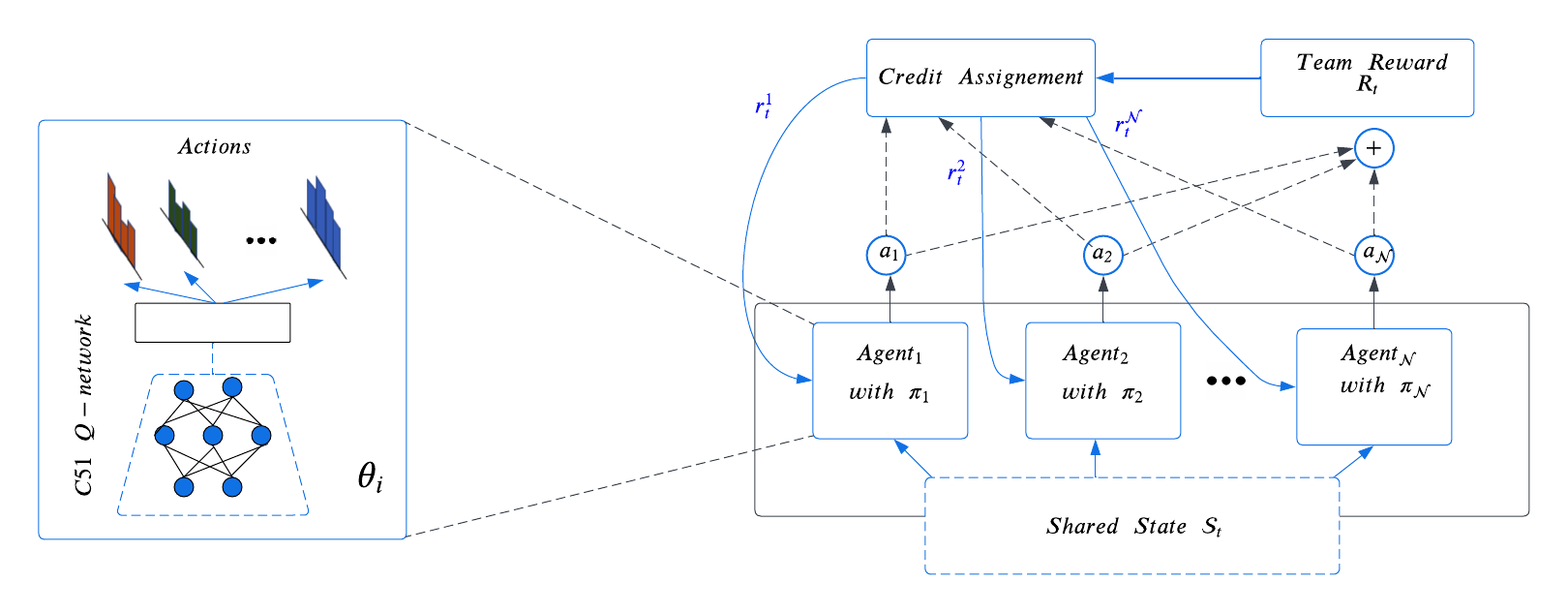}
\caption{Agent networks ad reward structure of the proposed MARL framework.}
\label{fig:onling learning}
\end{figure*}

\subsection{MARL Learning Design}

In this section, we describe the overall architecture of our path planning framework illustrated in Figure \ref{fig:onling learning}. As may be seen, an unconventional independent learning scheme is adopted, where agents are trained separately based on their respective actions and rewards. Doing so reduces considerably the action space of the markov game. Furthermore, the state space dimension is also reduced by directly handling the drones' budgets at the environment simulation level. To encourage cooperation, the global state is shared among agents, and each drone's contribution is deduced from the team reward, following its local action. Formally, a credit assignment strategy is defined as a function $\eta : \mathcal{S} \times \mathcal{A} \times \mathcal{R}  \rightarrow \mathbb{R}^{N}$ that
maps $\{ s_{t}, a_{t}, R_{t} \} $ to $[r^{u_{1}}_{t}, ..., r^{u_{N}}_{t}]$. Here, we consider two different
strategies of $ \eta (s_{t}, a_{t}, R_{t})$ for our MARL framework. The first strategy considers an equal split of the team reward $R_{t}$ among drones regardless of their actions, except for the drones that have completed, which receive a zero reward. For the second approach, we design a credit assignment strategy based on difference rewards. It captures an agent's contribution considering the difference rewards gain when computing rewards using the joint action and when excluding an agent specific action. Equation \ref{eq:diff_reward_c} gives the reward's expression for a given drone $u_{i} \in \mathcal{U}$. 

\begin{align}
\begin{split}
 r^{u_{i}}_{t}& = R_{t}. c^{u_{i}}_{t}
  \\
c^{u_{i}}_{t} &= \frac{1}{N} \underbrace{ - \frac{1}{\gamma} \frac{r(s_{t}, a^{\Bar{u_{i}}}_{t})-  \mathbb{E}[r(s_{t}, a^{\bar{u_{k}}}_{t}))] }   {\max \{r(s_{t}, a^{\bar{u_{k}}}_{t})\}- \min \{r(s_{t}, a^{\bar{u_{k}}}_{t})\}   } }_{\varepsilon_{i}} 
\end{split}
\label{eq:diff_reward_c}
\end{align}

where $c^{u_{i}}_{t}$ represents the contribution percentage of agent $u_{i}$ to the team reward $R_{t}$. This rate is calculated using the gain obtained by the teammates when excluding the action of agent $u_{i}$, denoted by $ r(s_{t}, a^{\bar{{u_{i}}}}_{t})$. 

Overall, each agent has at the beginning the same contribution $\frac{1}{N}$ that is adjusted following its action by the term $\varepsilon_{i}$, where $\gamma$ is a parameter controlling the impact of the actions taken, by allowing bigger or smaller variances between agents contributions.

In this work, we adopt the categorical Deep Q-Network (DQN) algorithm, called C51, a recent version of the DQN approach, that considers distributional returns reported as more stable during training \cite{bellemare2017distributional}. All our agents have the same Q-network architecture consisting of two fully connected hidden layers with 128 and 64 units, that have a shared state input. One state space sample contains drones' coordinates at time step $t$ and the actual estimated pollution map (output of the data assimilation after UAVs movements) and each Q-network outputs the probability distribution of the $|\mathcal{A}^{u} |$ possible actions. During training, we select the $\varepsilon$-greedy strategy for exploration and we consider episode termination when budgets of all UAVs are consumed.  Furthermore, to make our MARL generalizable to different starting locations, we consider random new starting sensing positions at each episode during training. Later this process can be extended to different starting budgets and different initial simulations.

\section{Air Pollution Considerations} \label{sec:pollution_considerations}
Knowing the limited drones flying time mission, we focus in this work on the steady dispersion case where concentrations values don't significantly change in time during drone mission. We leave the study of the unsteady dispersion for future work.
Without loss of generality, we focus on the best linear unbiased estimator, one of the most used data assimilation methods in the literature \cite{kalnay2003atmospheric}. 

To model BLUE parameters, we assume that sensors' measurements are uncorrelated, as they are linked to the electronic mechanism of sensors. Therefore, the covariance matrix of observation errors $R \in \mathbb{R}^{m \times m}$ is a diagonal matrix given by:
\begin{equation}
    R = v_{0}I
\end{equation}
where $v_{0}$ represents the observational error variance.
On the other hand, knowing that air pollution concentrations are spatially correlated, the simulation errors covariances can be expressed in function of the correlation coefficient as follows:

\begin{equation}
    cov(\epsilon^{b}_{i},\epsilon^{b}_{j})= w_{ij}\sqrt{\sigma^{2}(\epsilon_{i}^{b})\sigma^{2}(\epsilon_{j}^{b})}
\end{equation}

with $\sigma^{2}(\epsilon_{i}^{b})$ and $\sigma^{2}(\epsilon_{j}^{b})$, representing the variances of simulations errors for the location $i$ and $j$, and $ w_{ij}$ is the correlation coefficient. Several works from the fluid mechanic field study the correlation coefficient estimation of pollution simulation models, and one known method is to estimate this coefficient as exponential distance-decay function \cite{tombette2009pm,tilloy2013blue,boubrima2019deployment}, given by :
\begin{equation}
    w_{ij} = e^{ - \delta d_{ij} }
\end{equation}

where $\delta$ corresponds to the attenuation factor and $ d_{ij}$ represents the euclidean distance between the spatial points $i$ and $j$. 
Last, without loss of generality, we assume that the standard deviation of simulations errors  $\sigma(\epsilon^{b})$ is linearly dependent on simulation values, as done in some previous literature work \cite{fekih2020regression}. The resulting line slope coefficient, which we denote $\alpha$, reflects and controls the quality of the simulations. Higher values of  $\alpha$ indicate poorer simulations, and conversely, lower values suggest better simulations. It's important to note that alternative methods for estimating simulation error variances could also be employed to evaluate our proposed approach.

\section{Experiments}
\subsection{Experimental Setup}
\subsubsection{Dataset}
The proposed MARL approach is evaluated on the Fusion Field Trial 2007 (FFT07) dataset, a high-resolution real-world dataset recorded by the U.S. Army’s Dugway Proving Ground (DPG), in Utah \cite{storwold2007detailed}. In the latter, 100 digital photoionization detectors measuring propylene were placed on a square grid terrain, spaced evenly at $50 m$ apart and $2 m$ above the ground, and a propylene gas tracer was released from multiple locations at $2 m$ from the ground, at a constant flow rate. Several real experiments were conducted, varying the number of sources from one to four, which led to several trials. It's worth mentioning that this dataset is one of the few real-world datasets available that enable the study of air pollution monitoring at such a scale.
In this paper, we mainly present the results of trial 28 having 3 sources. Similar results can be found with other trials but the latter are not provided here due to page limit restrictions.

\subsubsection{Implementation and Hyperparameters}
A squared-shape geographical area of $450 m \times 475 m$ is selected for experiments. This region is organized in $10 \times 10$ sized grid, allowing 100 possibles moves for UAVs $(|\mathcal{A}^{u}|=100)$. The ground truth map corresponding to this monitored region is retrieved from trial 28 of the real-world dataset FFT07 (obtained from the fourth minute from the beginning of the release where the pollution plume is steady). Pollution concentrations across the map ranged from 0 to 300 part per million (ppm) with a standard deviation of 34 ppm. From the ground truth map, several simulations were obtained and averaged. These simulations are assumed to have normally distributed errors with a mean of zero and a covariance matrix, denoted by $B$, constructed following the same process as in section \ref{sec:pollution_considerations}. Furthermore, the variance of observation errors $v_{0}$ is considered negligible in this work. We leave the study of the impact of the latter on the mapping quality for future works.

To train RL agents, a simulator is developed in Python and used as the environment. Moreover, C51 neural networks are implemented using TF agents, a TensorFlow library for RL algorithms. Adam \cite{kingma2014adam} is utilized as the
optimizer for parameter estimation.
Table \ref{tab:rl_params} provides a summary of the values of the main parameters used in this work. To evaluate the performance of the compared algorithms, we used the metric of Mean Absolute Error (MAE). Later, all results of this work were obtained by averaging several independent runs of the training algorithm. 

\begin{table}
\centering
\caption{MARL experiments settings.}
\begin{tabular}{ll}
\toprule
Parameter                       & Value              \\ \midrule
Grid size                       &   $100$          \\ 
Number of drones   $N$             &   $ 1, 2, 3$             \\ 
Drones Budgets   $B_{max}$                      & $1000$ $m$,   $2000$ $m$,    \\ 
                    &   $3000$ $m$, $4000$ $m$ \\ 
Simulation errors coefficient $\alpha$             & $ 0.1, 0.3, 0.5, 0.7$    
\\ 
Factor used in equation \ref{eq:diff_reward_c}              & $\gamma= \frac{1}{N}$ \\ 
\midrule
\midrule

Replay buffer size              & $3e5$                  \\ 
Discount factor gamma           &   $0.99$               \\ 
Target network update frequency & $3$              \\ 
Minibatch size                  &     $128$                \\ 
Learning rate                   & $1e-5$   \\
\bottomrule
\end{tabular}

\label{tab:rl_params}
\end{table}

\subsubsection{Baselines}
To provide a meaningful comparison, we implemented a theoretical MARL path planning strategy that rely heavily on ground truth data. We define its team reward as follows:

\begin{equation}
R_{t} = 1 - \frac{MAE(x^{t}, x^{a}_{t})}{MAE(x^{t}, x^{b})}
\label{eq:reward_gt}
\end{equation}

The design of this equation indicates that higher (smaller) estimated errors give less (more) reward for the actions undertaken. Furthermore, we consider as for our MARL framework, two credit assignment strategies, namely equal reward split and difference rewards. We define the mathematical equation of individual agents' rewards for difference rewards method as follows :

\begin{align}
\label{eq:reward_dif}
\begin{split}
 r^{u_{i}}_{t}& = R_{t}. c^{u_{i}}_{t}
  \\
c^{u_{i}}_{t} &= \frac{1}{N-1}(1-\frac{r(s_{t}, a^{\Bar{u_{i}}}_{t})}{\sum_{u_{j \in \mathcal{U}}} r(s_{t}, a^{\Bar{u_{j}}}_{t})})
\end{split}
\end{align}
Here again, the agent's contribution $c^{u_{i}}_{t}$
is deduced from the total reward by considering the gain obtained by the teammates when excluding the action of agent $u_{i}$, denoted by $r(s_{t}, a^{\Bar{u_{i}}}_{t})$. $c^{u_{i}}_{t}$ is normalized so that $\sum c^{u_{i}}_{t}$ is equal to 1.  Although more informative than our MARL approaches, since the agents are guided by real errors, these MARL variants nevertheless remain unrealistic in practice given that the agents' learning and guidance are performed using ground truth data in all locations.  

Furthermore, we consider as another baseline a random navigation heuristic. The latter considers an initial aleatory drones deployment and then selects randomly at each step the new drones positions from the pool of the available actions so that UAVs don't move to the same locations. Although simple, random navigation has been often used as a baseline for many path planning problems \cite{park2022cooperative}. 

\begin{table*}
\centering
\caption{MAE obtained using 3 agents, considering 4 values of the simulation coefficient $\alpha$. All the results are presented in a 95\% confidence interval.}
\begin{tabular}{lllll}
\toprule
\multirow{2}{*}{Method} & \multicolumn{4}{c}{Performance (ppm)} \\
\cmidrule(lr){2-5} 
 & $\alpha = 0.1$ & $\alpha = 0.3$ & $\alpha = 0.5$ & $\alpha = 0.7$ \\
\midrule
Initial mean simulation error & $1.111$ & $2.154$ & $5.077$ & $8.072$ \\
DA-IQL-DiffRewards & $0.119 \pm 0.047$ & $0.139 \pm 0.020$ & $0.522 \pm 0.495$ & $0.615 \pm 0.236$ \\
DA-IQL-EqualSplit  & $0.193 \pm 0.127$ & $0.291 \pm 0.187$ & $0.987 \pm 0.927$ & $1.166 \pm 0.120$ \\
GT-IQL-DiffRewards  & $0.079 \pm 0.032$ & $0.132 \pm 0.023$ & $0.452 \pm 0.177$ & $0.484 \pm 0.163$ \\
GT-IQL-EqualSplit  & $0.083 \pm 0.030$ & $0.135 \pm 0.007$ & $0.558 \pm 0.139$ & $0.492 \pm 0.287$ \\
Random Navigation  & $0.322 \pm 0.105$ & $1.283 \pm 0.235$ & $2.420 \pm 1.195$ & $5.945 \pm 1.841$ \\
\bottomrule
\end{tabular}
\label{tab:diff_simulations}
\end{table*}

\subsection{Path Planning Performance}
For the sake of simplicity, in all the following evaluations, we refer to our proposed MARL solutions which are based on Data Assimilation errors as DA-IQL-DiffRewards and DA-IQL-EqualSplit, depending on the credit assignment strategy used (difference rewards and equal split rewards). Similarly, baselines using Ground Truth data are denoted GT-IQL-DiffRewards and GT-IQL-EqualSplit.

\begin{figure}[h]
\begin{center}
\includegraphics[scale=0.5]{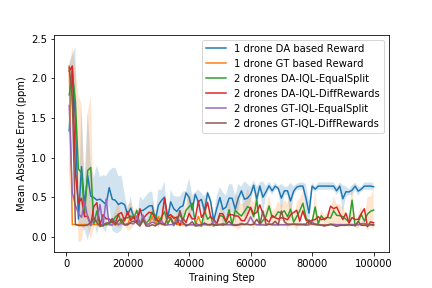}
\end{center}

\caption{The episodic mean ± std of one's and two agents' total mean absolute error for 5 random runs over $1e5$ training steps, with the evaluation performed every 1000 steps.}
\label{fig:training}
\end{figure}

The convergence curves of mean episodic total MAE of 1 and 2 agents by the different RL based approaches are illustrated in Figure \ref{fig:training}. In this scheme, the drones flying budget is set to $2000$ $m$ and the initial simulated map presents a simulation error coefficient $\alpha$ of $0.3$. From the resulting plot, we observe that all considered policies converge to relatively small estimation errors whatever the number of drones and the credit assignment strategy considered. This result is expected for one-drone approaches as the convergence of single-agent Q-learning is guaranteed in MDP domains. Furthermore, it's worth noticing that for approaches considering mean speed DA convergence as a reward, the increase in the number of drones, leads to better scores, while for theoretical policies based on ground truth, the impact of going from 1 to 2 agents, doesn't significantly impact performances. This is mainly due to the use of a more informative reward based on all ground truth data. Last, as expected, theoretical approaches exhibit better MAE with lower variances than MARL techniques using mean speed DA convergence as rewards. Nonetheless, the DA-IQL-DiffRewards attains very close performances to GT-based strategies.

\begin{figure}[h]
\begin{center}
\includegraphics[scale=0.5]{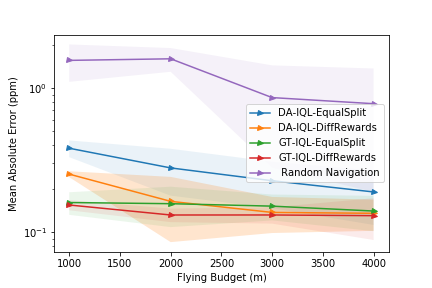}
\end{center}

\caption{MAE obtained using 2 agents considering different budgets for 5 random runs. Results are presented on a logarithmic scale.}
\label{fig:diff_budgets}
\end{figure}

Next, we present the results of the different approaches considering various drones' initial budgets and the same settings as previously in Figure \ref{fig:diff_budgets}. From this figure, it can be seen that for all considered methods, the increase in the flying budget, results in a decrease in the estimation error and an improvement of the assimilated pollution map. Among all path planning methods, the random navigation method shows the weakest scores whatever the flying budget.
Furthermore, we notice that DA-IQL-DiffRewards method shows better performance than DA-IQL-EqualSplit in all scenarios, reaching an estimation error up to $40\%$ lower. This is also observed for MARL methods based on ground truth where the difference-rewards based approach achieves $9\%$ better on average than the equal split MARL method. Also, it's worth noting that DA-IQL-DiffRewards achieves slightly better performance than the theoretical MARL GT-IQL-EqualSplit for high budgets and show close scores as GT-IQL-DiffRewards. This demonstrates that our reward shaping considering the average speed of DA convergence, combined with the proposed difference rewards splitting strategy, allows good air pollution mapping without having access to ground truth data.   

Last, to investigate the impact of the simulation error coefficient on the mapping quality, we vary $\alpha$ as follows $\alpha \in \{0.1, 0.3, 0.5, 0.7\}$ and we consider 3 drones having each a budget of $2000$ $m$. Table \ref{tab:diff_simulations} shows the obtained MAE for all considered approaches. The experimental results exhibit the following findings. i) All methods are impacted by the quality of the initial simulated map, showing most of the time larger estimation errors for larger $\alpha$ values. However, MARL methods are way less sensitive to this parameter than random navigation, when considering the initial mean simulation error. 
ii) When going from reliable simulations to less reliable ones ($\alpha =0.1$ to $\alpha =0.7$), the mapping quality of our DA-IQL-DiffRewards is at each time quite close to theoretical approaches, and the increase in estimation errors is not significant relative to initial mean simulation errors. This demonstrates that this strategy integrates quite well simulations errors and hence provides good pollution estimates regardless of the quality of the initial background.

\subsection{Scalability study}
While larger drone teams might be advantageous in vast environments, for our specific case (a $475m \times 450m$ area), adding more drones yielded minimal improvements in mapping quality. Therefore, we present results for up to three-drones configuration for clarity and focus. Nonetheless, we provide complexity analysis for our solution. The time complexity in the inference phase can be expressed as $O(d * N * (|\theta| + |A| + |S|))$ $ + O(d * $DA\_complexity$)$, with $d$ representing the episode length, $N$ the number of drones and $\theta$ the network parameters. Data assimilation complexity depends on the specific method and its implementation. For our case, it can be expressed by $O(|A|^{3}$) in worse cases. Overall, each step involves iterating over agents, executing a neural network forward pass to estimate action-values, and selecting the action with the highest Q-value. Data assimilation follows. The obtained results support the scalability of our solution with respect to the number of drones.

\section{Conclusion}

In this work, we study the UAVs' trajectories planning problem for air pollution mapping through data assimilation and we propose a cooperative multi-agent reinforcement learning. In the designed approach, drones agent collaborate to ensure the improvement of data assimilation outputs, while relying on a ground truth data-free reward, sharing their states and splitting their team rewards following two credit assignment strategies. Experiments using real-world data show that our difference-rewards based MARL achieves comparable results to a theoretical ground truth based MARL in terms of estimation errors. We also presented the effectiveness of our proposal for various flying budgets and different quality simulated maps including low drones budgets and poor initial simulations. Last, the rapid convergence exhibited by our MARL model suggests its possible adaption to unsteady pollution cases by considering an online update of our model.

\bibliographystyle{IEEEtran}
\bibliography{irosbib}


\end{document}